\documentclass{article}
\usepackage{spconf,amsmath,graphicx}

\usepackage{iftex}
\usepackage{mathptmx}
\ifPDFTeX
\else
  \usepackage{fontspec}
\fi
\usepackage{booktabs}
\usepackage{enumitem}
\setlist{nosep, leftmargin=12pt}
\usepackage[hidelinks]{hyperref}
\graphicspath{{figures/}{../out/paper/}}

\title{TuMAP: A Dual-View, Morphotype-Aware Pathomics Pipeline for Quantifying Renal Tubular Remodeling}

\name{
\begin{tabular}{c}
Ruijie Wu$^{1}$\qquad
Yuechen Yang$^{1}$\qquad 
Yanfan Zhu$^{1}$\qquad 
Yu Wang$^{2}$\qquad
Shilin Zhao$^{2}$\qquad \\
Ming-Tsun Tsai$^{3,4}$\qquad
Szu-Yuan Li$^{3,4}$\qquad
Shuo-Ming Ou$^{3,4}$\qquad
Tz-Heng Chen$^{3,4}$\qquad \\
Haichun Yang$^{2}$\qquad 
Yuankai Huo$^{1,{\star}}$\thanks{Email: \href{email:yuankai.huo@vanderbilt.edu}{yuankai.huo@vanderbilt.edu}}
\end{tabular}
}

\address{
$^{1}$ Vanderbilt University, Tennessee, USA \\
$^{2}$ Vanderbilt University Medical Center, Tennessee, USA \\
$^{3}$ Division of Nephrology, Department of Medicine, Taipei Veterans General Hospital, Taipei, Taiwan \\
$^{4}$ School of Medicine, College of Medicine, National Yang Ming Chiao Tung University, Taipei, Taiwan
}

\begin{document}
\maketitle

\begin{abstract}
Quantifying disease-associated renal tubular remodeling is challenging because tubular appearances are heterogeneous. We introduce TuMAP (Tubular Morphotype-Aware Pathomics), a pipeline combining (1) dual-view foundation-model representations and provisional morphotype annotation, (2) compositional and within-morphotype pathomics, and (3) patient-level statistical inference. From 456,158 tubules in 358 PAS-stained kidney biopsy slides, TuMAP identified eight image-based morphotypes, seven of which showed stable subsampling assignments. Across 219 patients in six diagnostic groups, diabetic nephropathy (DN) and tubulointerstitial nephritis/acute tubular injury (TIN/ATI) each differed from minimal change disease in seven morphotype proportions, while IgA nephropathy differed in six. DN showed reduced nuclear solidity and increased eccentricity across all eight morphotypes. Area-normalized nuclear density exposed additional within-morphotype differences. TuMAP provides a framework for separating changes in tubular morphotype composition from phenotypic differences within morphotypes.

\end{abstract}

\begin{keywords}
Computational Pathology, Renal Tubular Morphology, Quantitative Pathomics, Patient-Level Inference
\end{keywords}

\begin{figure*}[t]
\centering
\includegraphics[width=0.82\textwidth]{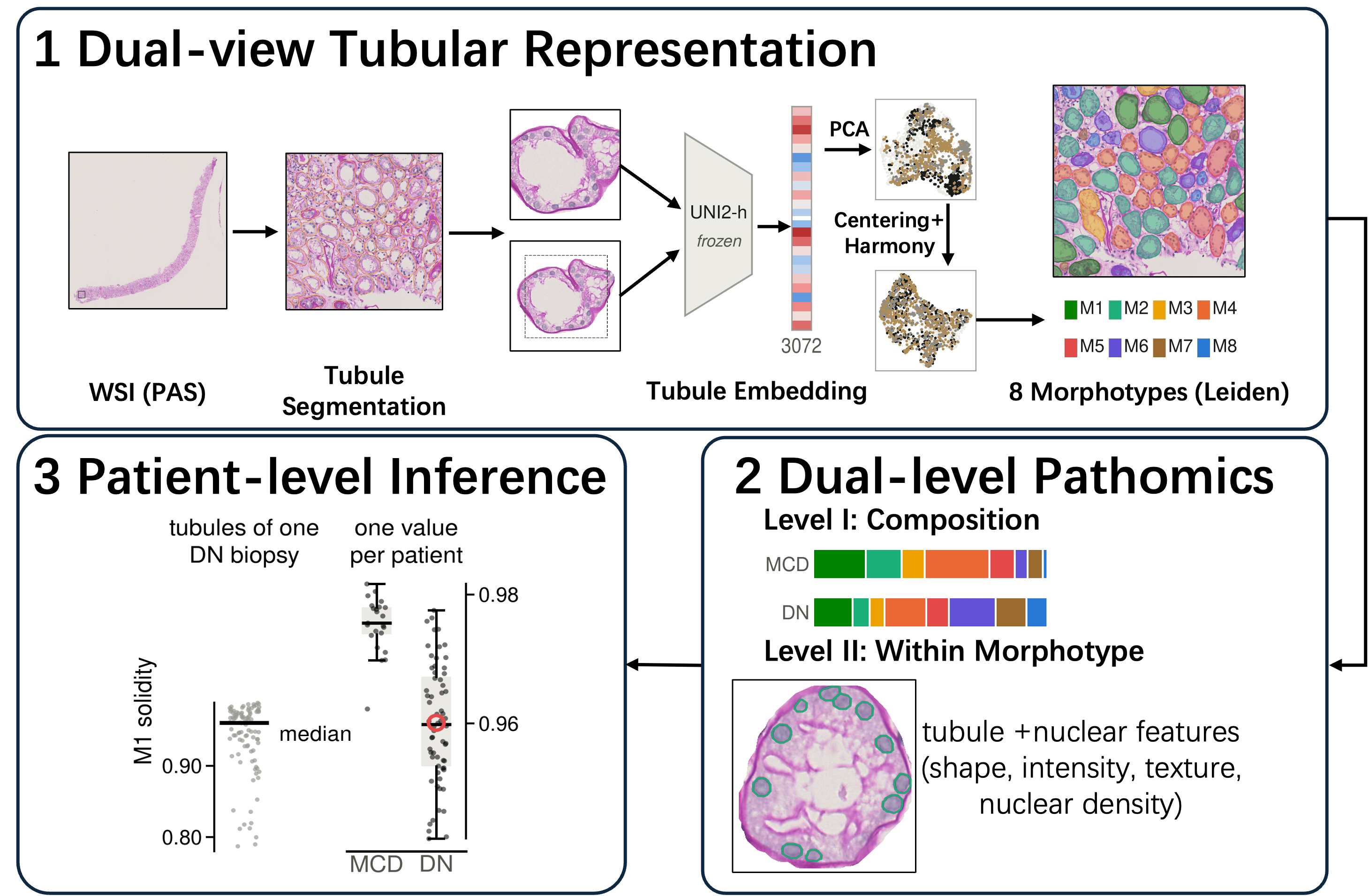}
\caption{\textbf{Overview of the TuMAP pipeline.} (1) \textbf{Dual-view tubular representation:} Texture and shape views encoded with frozen UNI2-h are harmonized and clustered into provisional M1--M8 morphotypes. (2) \textbf{Dual-level morphotype-aware pathomics:} Composition and within-morphotype phenotypes are quantified. (3) \textbf{Patient-level statistical inference:} Measurements are aggregated per patient and morphotype (red circle: M1 median solidity in one DN biopsy; note different y-axis scales), followed by regression with HC3 robust standard
errors and FDR control.}
\label{fig}
\label{fig:workflow}
\end{figure*}

\section{Introduction}
\label{sec:intro}

Renal tubular remodeling is a major manifestation of kidney disease, yet its histopathological assessment remains largely semi-quantitative~\cite{barisoni2020digital}. Recent computational pathology methods enable large-scale renal structure segmentation and morphometry~\cite{lucarelli2023reference,holscher2023flash}, including patient-level tubular analysis~\cite{fan2025tubular}, individual tubule characterization~\cite{kavthekar2022tubular}, and histological phenotype discovery~\cite{pandit2025ssl,joodaki2024pilot}. However, an integrated quantitative pipeline for systematically characterizing disease-associated tubular remodeling remains lacking.

Three challenges remain: representing texture and shape despite slide variation; separating changes in morphotype composition from differences within morphotypes; and avoiding pseudoreplication when inferring disease associations from correlated tubules~\cite{squair2021confronting,zimmerman2021pseudoreplication}.

We present \textbf{TuMAP (Tubular Morphotype-Aware Pathomics)}, a framework linking object-level morphological characterization to patient-level inference, inspired by single-cell differential abundance and pseudobulk analysis~\cite{phipson2022propeller}. By examining morphotype composition and interpretable phenotypes within morphotypes separately, TuMAP assesses whether disease-associated differences involve changes in the mixture of tubular appearances, alterations within broadly similar groups, or both. Our contributions are:

\begin{itemize}
    \item \textbf{Dual-View Tubular Representation:} Integrating foundation model-derived texture and shape embeddings with cross-slide harmonization for unsupervised morphotype discovery.

    \item \textbf{Dual-Level Morphotype-Aware Pathomics:} Quantifying both disease-associated morphotype compositional shifts and within-morphotype tubular and nuclear morphological alterations.

    \item \textbf{Patient-Level Statistical Inference:} Combining patient-level aggregation, covariate-adjusted regression, and sensitivity analyses for disease association analysis while avoiding pseudoreplication.
\end{itemize}

\begin{table}[t]
\centering
\caption{\textbf{Comparison cohort (219 patients).} Tubules: total on analyzed sections; tubules/patient: median [IQR]. GS: globally sclerosed glomeruli (\%), median [IQR] (190 patients with available GS\%).}
\label{tab:cohort}
\footnotesize
\setlength{\tabcolsep}{3.5pt}
\begin{tabular}{lrrrr}
\toprule
Diagnosis & Patients & Tubules & Tubules/patient & GS (\%) \\
\midrule
MCD & 23 & 33,486 & 1446 [1156-1722] & 4.3 [0.0--13.4] \\
MN & 37 & 53,603 & 1426 [1154-1714] & 4.6 [0.0--15.6] \\
LN & 28 & 43,628 & 1418 [1197-1980] & 9.4 [2.9--23.6] \\
IgAN & 39 & 43,228 & 1049 [879-1234] & 22.2 [10.0--50.0] \\
TIN/ATI & 28 & 30,931 & 1070 [805-1402] & 20.0 [0.0--34.4] \\
DN & 64 & 78,236 & 1164 [1008-1388] & 52.7 [33.3--64.6] \\
\midrule
Total & 219 & 283,112 & 1253 &  \\
\bottomrule
\end{tabular}
}
\end{table}

\begin{figure*}[t]
\centering
\includegraphics[width=0.90\textwidth]{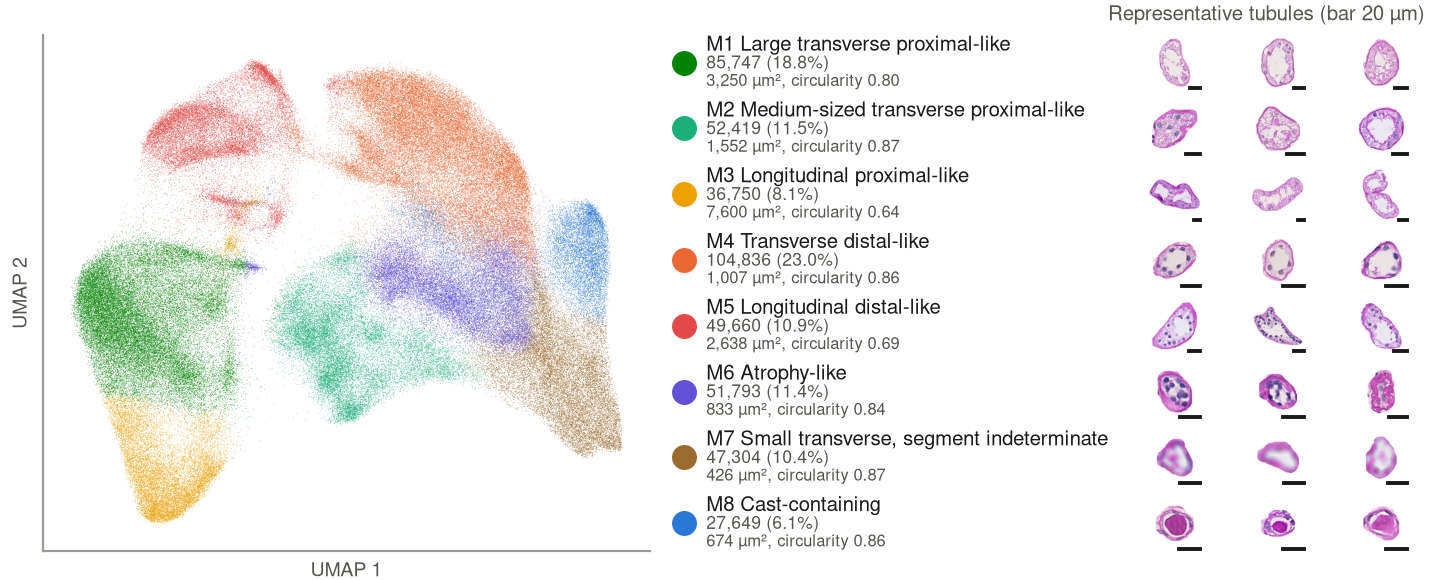}
\caption{\textbf{Eight provisional morphotypes (456,158 tubules; 358 slides).} Left: UMAP (200,000 points). Right: M1--M8 counts, median area/circularity, and representative tubules (20-\textmu m bars). Segment-like names are not immunohistochemically confirmed.}

\label{fig:umap}
\end{figure*}

\begin{figure*}[t]
\centering
\includegraphics[width=\textwidth]{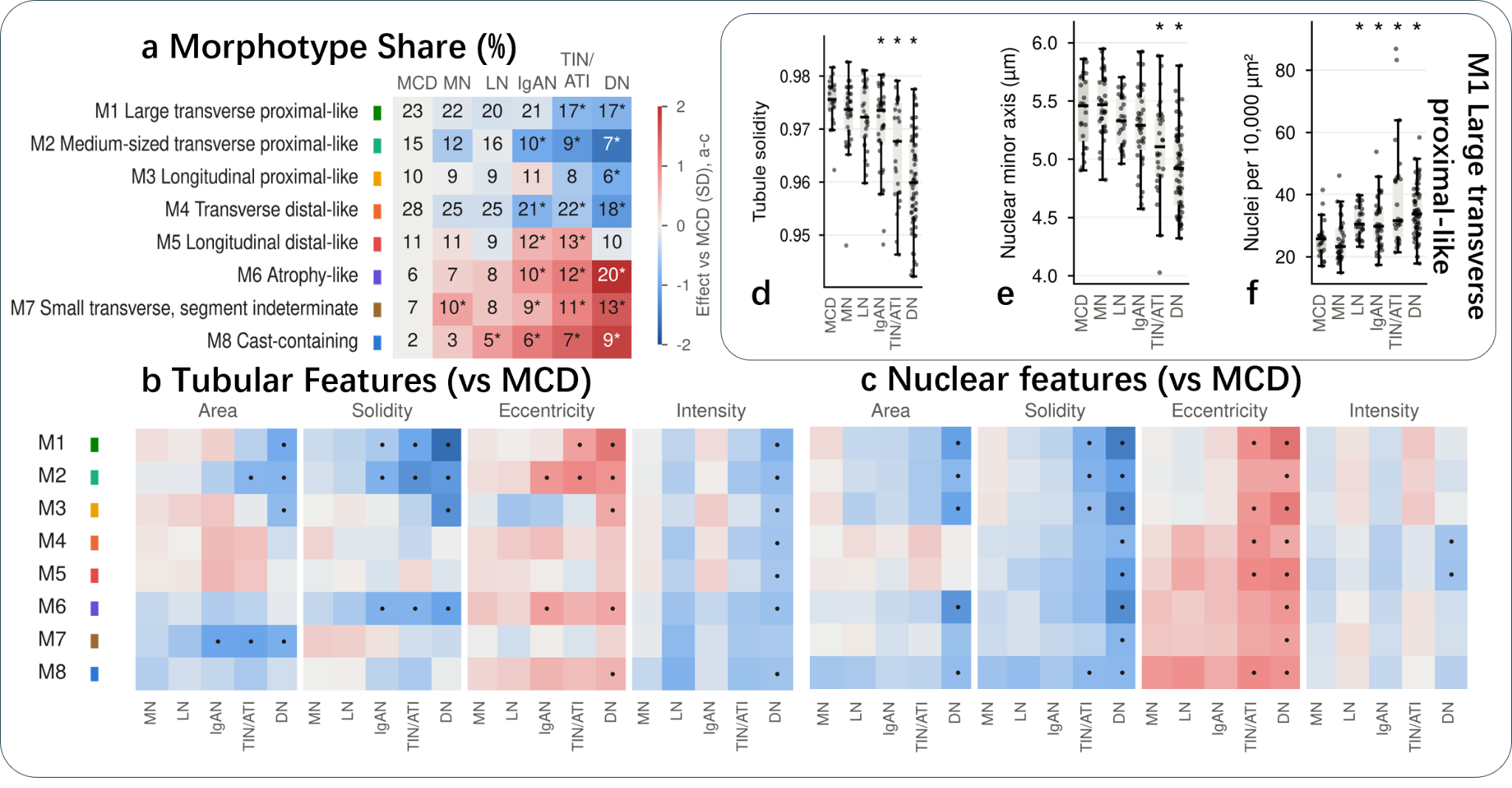}

\caption{\textbf{Patient-level comparisons vs.\ MCD ($n=219$).} (a) Mean morphotype shares (\%) with adjusted logit-share effects (SD). (b,c) Tubular/nuclear effects (SD), labeled by M number; dots mark FDR $q<0.05$. (d--f) M1 tubular solidity, nuclear minor axis, and assigned nuclei per 10,000\,\textmu m$^2$ of contour area (including lumen). Points represent patients; asterisks mark FDR $q<0.05$.}

\label{fig:results}
\end{figure*}

\section{Methods}
\label{sec:methods}

\subsection{Study Cohort and Data Preparation}

The Taipei Renal Transcriptomics and Outcomes Investigation
(TRTOI), a prospective cohort established in 2018, provided
363 PAS-stained native kidney biopsy slides from 362 patients
(0.221--0.230~\textmu m/pixel). After quality control, 358 slides/357 patients entered diagnosis-agnostic discovery; 351 had pathology reports. Patients were grouped by unambiguous primary diagnosis; accompanying tubular, interstitial or vascular findings were permitted, but additional glomerular diseases excluded cases from MCD, DN, MN, LN and IgAN groups. The comparison comprised 219 patients with diabetic nephropathy (DN), IgA nephropathy (IgAN), membranous nephropathy (MN), lupus nephritis (LN), minimal change disease (MCD), or tubulointerstitial nephritis/acute tubular injury (TIN/ATI); 132 others contributed only to discovery (Table~\ref{tab:cohort}). MCD was a diseased reference; eight of 23 MCD patients had accompanying ATI/TIN. GS\% required at least five manually annotated glomeruli.

\subsection{Dual-View Tubular Representation}
\label{sec:representation}

An in-house pipeline based on YOLOv12 and SAM 2~\cite{yolov12_ultralytics,ravi2025sam} segmented tubules. Texture and contour-masked shape crops ($224^2$ pixels; 0.45\,\textmu m/pixel) were encoded by frozen UNI2-h~\cite{chen2024uni}; the concatenated 3,072-D embeddings were $\ell_2$-normalized, reduced by PCA to 50 dimensions, slide-centered, and integrated with Harmony~\cite{korsunsky2019harmony}. Leiden clustering~\cite{traag2019leiden} used 25 nearest neighbors (resolution 0.5). Stability was tested by 80\% subsampling (ARI/Jaccard). In an 11-slide/42,827-tubule pilot, dual views improved slide-held-out $R^2$ for major axis (0.83 vs.\ 0.74/0.82) and Haralick texture (0.62 vs.\ 0.60/0.57 for texture/shape alone).

\noindent\textit{Provisional morphotype annotation.} M1--M8 were annotated along three descriptive axes: segment-like identity (proximal-like, distal-like, or indeterminate), sectioning plane (transverse, longitudinal, oblique, or indeterminate), and lesion pattern (including atrophy-like change, dilation, and casts). The designation \emph{no overt lesion} does not imply normality; sizes distinguish otherwise similar appearances. Uncertain axes remained indeterminate, and labels describe
predominant rather than universal cluster features
(Fig.~\ref{fig:umap}).

\subsection{Dual-Level Morphotype-Aware Pathomics}
\label{sec:pathomics}

TuMAP analyzes two levels: \textbf{Level I: Morphotype composition}, the patient-level morphotype proportions, and \textbf{Level II: Morphotype-conditioned morphology}, tubular and nuclear features within each morphotype.

PySpatial~\cite{yang2025pyspatial} yielded 119 tubular shape, intensity, Haralick texture, and radial features, excluding rotation/reflection-sensitive features (ICC $<0.75$). Nuclei were segmented using a Cellpose-SAM~\cite{pachitariu2025cellposesam}
checkpoint fine-tuned on expert-annotated challenging kidney nuclei
in our prior work~\cite{wu2026renal_nuclei}.
Nuclei were assigned to tubules by centroid inclusion
within contours dilated by 2\,\textmu m. We extracted 97 nuclear features after Macenko stain deconvolution~\cite{macenko2009normalizing}, excluding nuclei deeper than 15\,\textmu m inside original contours for nuclear morphology. Separately, \textit{nuclear density} was defined per tubule as all assigned nuclei divided by original contour area (including lumen), expressed per 10,000\,\textmu m$^2$, without the 15-\textmu m depth filter.

\subsection{Patient-Level Statistical Inference}
\label{sec:inference}

To avoid pseudoreplication, TuMAP aggregates measurements at the patient level using one slide per patient. Morphotype counts $x_{ik}$ among $N_i$ patient tubules were transformed using a continuity-corrected logit~\cite{phipson2022propeller}:
\[
z_{ik}=\operatorname{logit}
\left(\frac{x_{ik}+0.5}{N_i+1}\right).
\]
Within each morphotype, patient medians summarized tubular features ($\geq10$ tubules), nuclear features ($\geq30$ qualifying nuclei), and per-tubule nuclear density (without a minimum-tubule threshold).

Standardized patient-level outcomes were regressed on diagnosis (MCD reference), scan-resolution group, and acquisition year; composition additionally adjusted for annotated glomeruli per 1,000 tubules. Effects are in SD units with 95\% confidence intervals from HC3 robust standard errors~\cite{mackinnon1985hc}. Benjamini--Hochberg FDR ($q<0.05$) was controlled by diagnostic contrast within composition, tubular, nuclear, and nuclear-density feature families~\cite{benjamini1995fdr}. Within-morphotype tests included 208--219 patients, depending on available measurements.

Sensitivity analyses compared GS\%-adjusted and unadjusted models in the same 190 patients, restricted diagnoses, excluded DN with arterionephrosclerosis, and imposed a stricter tubule-detection confidence threshold ($\geq0.93$).

\section{Results}
\label{sec}

\hspace*{\parindent}\textbf{1. Unsupervised Tubular Morphotype Discovery.} Diagnosis-agnostic clustering of 456,158 tubules from 358 slides identified eight provisional image-derived morphotypes (Fig.~\ref{fig:umap}): proximal-like M1--M3, distal-like M4--M5, atrophy-like M6, small transverse M7, and cast-containing M8. Median areas spanned 426--7,600~$\mu\mathrm{m}^2$. Morphotypes were broadly distributed across slides (effective slide diversity 192--293; maximum single-slide share 2.1\%). Subsampling yielded ARI $0.75 \pm 0.07$ and Jaccard indices of 0.69--0.91 for seven morphotypes (M6: 0.51). M6--M8 proportions correlated positively with GS\% ($\rho=0.61$--$0.68$), while proximal-like proportions correlated negatively ($\rho=-0.29$ to $-0.58$).

\textbf{2. Disease-Associated Morphotype Composition.} Compared with MCD, DN and TIN/ATI each differed in seven of eight morphotype proportions, IgAN in six, and MN and LN in one each (M7 and M8, respectively; Fig.~\ref{fig:results}(a)). In DN, mean M6 share increased from 5.5\% to 20.1\% ($+2.00$ SD; 95\% CI, 1.63--2.37), M8 from 1.9\% to 9.0\% ($+1.78$ SD), and M7 from 6.6\% to 13.1\% ($+1.52$ SD). All three proximal-like shares declined, including M2 (15.3\% to 7.4\%; $-1.56$ SD).

\textbf{3. Within-Morphotype Morphological Differences.} Within morphotypes, DN nuclear solidity decreased (median $-1.06$ SD) and eccentricity increased ($+0.88$ SD), each significant in all eight morphotypes; TIN/ATI showed corresponding differences in four and five, respectively (Fig.~\ref{fig:results}(b--c)). DN tubular PAS intensity was lower in seven morphotypes. In proximal-like M1, predominantly annotated as lacking overt lesions, DN was associated with lower tubular solidity ($-1.65$ SD; 95\% CI, $-1.95$ to $-1.34$) and nuclear minor axis ($-1.32$ SD; 95\% CI, $-1.72$ to $-0.92$). M1 area-normalized nuclear density was higher in TIN/ATI ($+1.26$ SD), DN ($+0.81$ SD), IgAN ($+0.50$ SD), and LN ($+0.47$ SD; Fig.~\ref{fig:results}(d--f)). Because density uses total tubular contour area, including lumen, increases need not indicate more nuclei.

\textbf{4. Sensitivity analyses assess robustness.} In the main analysis, 373 of 952 tubular and
370 of 776 nuclear feature tests were
FDR-significant ($q<0.05$) for DN versus MCD,
compared with 107 and 145, respectively,
for TIN/ATI and 89 and 0 for IgAN;
none were significant for MN or LN
in either family. Diagnostic restrictions and exclusion of DN with arterionephrosclerosis preserved 92--100\% of effect magnitudes; a stricter detection threshold retained 70--87\% of DN tubular-solidity effects. Among the same 190 patients with GS\% data, adjustment retained median effect magnitudes of 45\% (DN nuclear), 59\% (DN tubular), and 84\% (TIN/ATI), suggesting overlap with GS\%-related pathological variation.

\section{Discussion}
\label{sec:conclusion}
We introduced TuMAP, a morphotype-aware quantitative pathomics pipeline integrating dual-view tubular representation, dual-level morphological profiling, and patient-level statistical inference. Applied to 456,158 tubules, TuMAP identified eight morphotypes and revealed disease-associated changes in both morphotype composition and within-morphotype tubular and nuclear morphology. Notably, DN and TIN/ATI exhibited widespread remodeling at both levels, while adjustment for glomerulosclerosis attenuated several associations, highlighting the importance of pathological severity in disease comparisons. These morphotypes represent provisional image-based groups rather than validated nephron segments, warranting further histological validation.

\section{COMPLIANCE WITH ETHICAL STANDARDS}
The TRTOI cohort study was approved by the Institutional
Review Board of Taipei Veterans General Hospital
(Approval No. 2018-06-008B).

\section{Acknowledgments}
\label{sec:acknowledgments}
No funding was received for conducting this study.
The authors have no relevant financial or non-financial
interests to disclose. ChatGPT was used to assist with language editing and proofreading.

\bibliographystyle{IEEEbib}
{\small\bibliography{refs}}   

\end{document}